\documentclass[10pt,conference]{ieeeconf} 
\IEEEoverridecommandlockouts                              

\usepackage{graphics} 
\usepackage{subfigure}
\usepackage{graphicx} 
\usepackage{epsfig} 
\usepackage{amsmath} 
\usepackage{amssymb}  
\usepackage{booktabs}
\usepackage{algorithm}
\usepackage{algpseudocode}
\usepackage{gensymb}
\usepackage{cite}
\usepackage{array}
\usepackage{multirow}
\makeatletter
\def\endthebibliography{%
	\def\@noitemerr{\@latex@warning{Empty `thebibliography' environment}}%
	\endlist
}
\makeatother

\usepackage{float}
\floatstyle{plaintop}
\restylefloat{table}

\usepackage{caption}
\usepackage{mathptmx}
\DeclareCaptionFont{lmss}{\fontsize{9}{11}\mdseries}
\usepackage[font=lmss,labelfont=bf]{caption}

\def\v#1{\mathbf{#1}}
\def\m#1{\mathsf{#1}}

\title{\LARGE \bf
Automated pick-up of suturing needles for robotic surgical assistance
}

\author{C. D'Ettorre, G. Dwyer, X. Du, F. Chadebecq, F. Vasconcelos , E. De Momi and D. Stoyanov
\thanks{C. D'Ettorre and E. De Momi are with the Department of Electronics, Information and Bioengineering, Politecnico di Milano, Milan, Italy {\tt\small  claudia.dettorre@mail.polimi.it , elena.demomi@polimi.it}} 
\thanks{G. Dwyer, X. Du,  F. Chadebecq, F. Vasconcelos and D. Stoyanov are with the Centre for Medical Image Computing (CMIC) and the Department of Computer Science, University College London, London, UK {\tt\small \{george.dwyer.14, xiaofei.du.13, f.chadebecq,f.vasconcelos, danail.stoyanov\}@ucl.ac.uk}}%
}

\begin{document}

\maketitle
\thispagestyle{empty}
\pagestyle{empty}

\begin{abstract}
  
Robot-assisted laparoscopic prostatectomy (RALP) is a treatment for prostate cancer that involves complete or nerve sparing removal prostate tissue that contains cancer. After removal the bladder neck is successively sutured directly with the urethra. The procedure is called urethrovesical anastomosis and is one of the most dexterity demanding tasks during RALP. Two suturing instruments and a pair of needles are used in combination to perform a running stitch during urethrovesical anastomosis. While robotic instruments provide enhanced dexterity to perform the anastomosis, it is still highly challenging and difficult to learn. In this paper, we presents a vision-guided needle grasping method for automatically grasping the needle that has been inserted into the patient prior to anastomosis. We aim to automatically grasp the suturing needle in a position that avoids hand-offs and immediately enables the start of suturing. The full grasping process can be broken down into: a needle detection algorithm; an approach phase where the surgical tool moves closer to the needle based on visual feedback; and a grasping phase through path planning based on observed surgical practice. Our experimental results show examples of successful autonomous grasping that has the potential to simplify and decrease the operational time in RALP by assisting a small component of urethrovesical anastomosis.    
    
\end{abstract}


\section{Introduction}

Robotic minimally invasive surgery (RMIS) is now an established alternative to open and laparoscopic surgery for the treatment prostate cancer. Robotic instruments offer increased dexterity and enhance surgical ergonomics through tele-operation, which helps surgeons to operate with less invasive approaches and results in a range of benefits for the patient like faster recovery time, less pain and reduced tissue trauma. As a result, robotic surgical platforms like the \textit{da Vinci} Surgical System by Intuitive Surgical (Sunnyvale, CA) are facilitating an increasing number of complex procedures to be performed through RMIS \cite{IntuitiveSurgical}. Despite the growing RMIS take-up, automation through the robotic platform is not currently available but if possible it could assist surgeons by helping and also standardising simple procedural tasks and even potentially removing errors \cite{alemzadeh2016adverse}.

\begin{figure}[!t]
	\centering
	\subfigure[\label{abc}]{\includegraphics[width=0.23\textwidth]{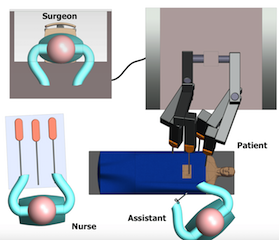}}
	\subfigure[\label{zzz}]{\includegraphics[width=0.23\textwidth]{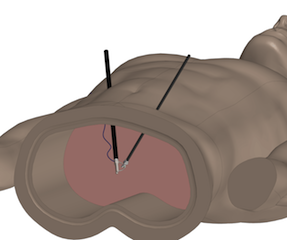}
	}
	\subfigure[\label{etc}]{
		\includegraphics[width=0.15\textwidth, angle =180]{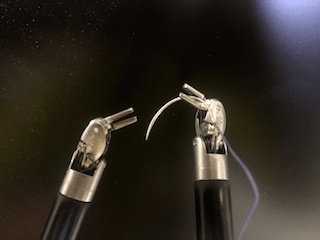}
		\includegraphics[width=0.15\textwidth, angle =180]{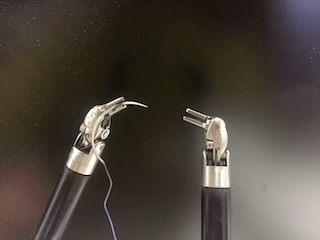}
		\includegraphics[width=0.15\textwidth, angle =180]{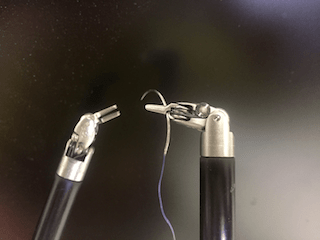}
	}
	\caption{ 
		\ref{abc}: Schematic view of the da Vinci Surgical System. The surgeon operates using a master console with tele-manipulated instruments in the patient. An assistant swaps and positions instruments into the surgical site (on the right) and inserts suturing needles. 
		\ref{etc}: Depiction of hand-offs phases showing the two Large Needle Driver (LND) tools. First, the right-hand tool grasps the needle at 2/3 of its length (starting from the tip), then it passes the needle to the left-hand tool that usually approaches at around 1/3 of the length. The last step is the final approach of the needle from the right-hand tool that need to grasp the needle in the proper final position. In case of vesicourethral anostomosis the same approach is then repeated for the other needle.} 
\end{figure}

In the case of Robot-assisted laparoscopic prostatectomy (RALP), the surgical operation and workflow consists of different steps that can be performed with some variation in their order, namely: lymph node and posterior dissection, incision and mobilisation of bladder and prostate, cancer excision, and after having completely removed the prostate, urethrovesical anastomosis \cite{higuchi2011robotic}. Each of these can in turn be broken down into sub-steps within a full surgical procedure ontology. Two main suturing techniques are used at the end of this procedure: interrupted suturing and running anastomosis (the Van Velthoven technique) \cite{van2003technique}. When using the da Vinci system, the dexterity of the endo-wrist technology helps significantly with suturing. Because the sutures can be placed at almost any angle, the running approach is normally preferred \cite{albisinni2016single}. Focusing on the phases of suturing during urethrovesical anastomosis, when the suturing phase starts, the surgical assistant introduces a circular needle inside the patient through a trocar using a needle grasper. As it is possible to see from Fig.\ref{abc} the assistant usually stands close to the insertion ports in order to cooperate with the surgeon. Then the surgeon first grasps the needle in the most comfortable configuration to insert it within the tissue and pass it between the tools, as shown in Fig.\ref{etc} \cite{zehtabchi2012impact}. Normally multiple hand-offs are required to optimise the needle position, since the manoeuvrability of the tool is fully surgeon controlled, and better robotic instrument joints configuration could practically be computed to simplify the grasping phase. There may be slight variation in the general hand-offs' execution according to the individual surgeon’s dexterity and experience which could be optimised to an agreed gold standard.

\begin{figure*}[!t]
	\centering
	\includegraphics[width=0.9\textwidth]{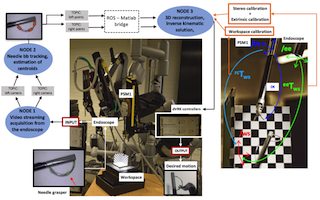}
	\caption{Overview of the system framework. For the experiment was used: 1/2 circular crested needle with green markers, single action laparoscopic needle grasper, a workspace characterised by three perpendicular checkerboard planes, chessboard 6cmx8cm with square dimension of 10 mm for extrinsic calibration and 3mm for stereo calibration. The relative position between the stereo endoscope, the grasper and the PSM was set so as to replicate as closely as possible the distances in a real surgical operation. Endoscope light intensity was set at 30/100. ROS architecture: blue ovals represent nodes, while grey squares topics. Transformation definition:\(\m{^{rc}T_{ws}}\) is the homogenous transformation matrix between the workspace reference frame \textit{(/ws)} and the endoscope one \textit{(/ee)}, while \(\m{^{ee}T_{ws}}\) represent the extrinsic pose of the left camera in respect of the \textit{/ws}.}
	\label{fgh}
\end{figure*}

In this paper, the aim of our work is to build the vision and robotic control algorithms required towards automation of needle grasping before robotic assisted suturing begins. We believe that optimal needle grasping can avoid hand-offs steps, reducing the amount of time required for vescicourethral anastomosis. Our approach includes an vision algorithm for needle tracking, a visual servoing system for the approaching phase and a needle grasping optimisation planning based on best practice of suturing in teleoperation mode. In all our experiments, we used one needle and one patient side manipulator (PSM) but in principle the procedures can be simply duplicated in order to use multiple PSMs or needles. We report  preliminary results of the calibration accuracy needed for our system to close the visual-servoing loop and we show promising qualitative demonstrations of needle grasping in practice.

Section II sets the background of the state-of-the art for surgical task automation. This is followed in section III by the definition of the problem, description of the system set-up and transformations between frames, control scheme and software architecture. The results of system performances, are shown in section V and the paper concludes with a discussion of findings and some planned future work.

%


\section{Background and Related Work}
RMIS is used for many abdominal tumorectomy interventions, such as prostatectomy, as described in reviews of recent developments in semi-autonomous and autonomous execution of surgical procedures by Moustris et al\cite{moustris2011evolution}. 

This work was developed using the da Vinci Research Kit (dVRK) platform, that is currently being used in 15 research labs for tasks ranging from autonomous tool tracking in ultrasound images \cite{mohareri2014vinci}, tissue palpation using an ultrasound probe for tumour detection \cite{billings2012system}, to multilater debridement and cutting tasks \cite{murali2015learning}.

\vspace{2mm}
\textbf{Automated Suturing - } Represents a well studied topic in the literature: Kang et al. \cite{kang2000autonomous} introduced a multi-step task planning based on hierarchical models, Schulman et al.\cite{schulman2013case} focused on the interaction with deformable tissue base on a non-rigid registration using a learning by demonstration approach, as previously done in \cite{jackson2012modeling}.
Chow et al. \cite{chow2014novel} proposed two autonomous knot-tying methods based on stereo vision. 
None of these works have properly faced the problem of needle grasping, even if it represents the starting point for all of them.
Collaborative human-robot suturing was shown by Padoy et al. \cite{padoy2011human}, although they required human interactions for needle insertion and hand-offs were performed manually.
There were many commercial efforts to mitigate back-and-forth hand-offs through passively orienting the needle on gripper closure using a “self-righting” gripper jaw design \cite{martin2015articulating}. However, these are not designed for automation, and require a complete tool redesign.
Staub et al.\cite{staub2010automation}, firstly analysed the needle alignment and tissue piercing, in order to automatise those phases, although they assumed the robot already holding the needle perpendicular to the jaws of the forceps. 
Recently, Siddarth et al. \cite{sen2016automating} worked on an automating multi-throw multilateral surgical suturing introducing a novel mechanical needle guide, and optimised the entire framework using sequential convex programming. They assessed needle grasping problem applied to the pulling phase during the suturing procedure. They developed a Suture Needle Angular Positioner aiming to reduce needle pose uncertainty, allowing higher tolerances in relative positioning but still maintaining hand-off procedure. 
Other surgical subtasks have been studied: multilateral debridement using Raven surgical Robot\cite{kehoe2014autonomous}, surgical cutting based on learning by observation (LBO) algorithm \cite{murali2015learning} and on deep reinforcement learning policies\cite{thananjeyan2017multilateral}.

\vspace{2mm}

\textbf{Visual Servoing - } It is a popular approach to guide a robotic tool using visual feedback from a camera system. P. Hynes et al. \cite{hynes2006uncalibrated},\cite{hynes2005uncalibrated} developed a robotic surgery system using visual servoing and conducted autonomous suturing. Their system is able to position the instrument and they performed suturing by setting the desired points manually. However, performance parameters of the system, such as positioning precision of the instruments, were not analysed.
Staub et al. \cite{staub2010autonomous} dealt with the automated positioning of surgical instruments by employing visual guidance.

\vspace{2mm}

This paper builds on a prior work based on a visual guidance \cite{nageotte2006visual} for the initial phase of needle localisation and approach, followed by a grasping motion definition built on a Finite State Machines (FSM) \cite{murali2015learning}.
To the best of our knowledge, we were unable to find any other related works treating the problem of grasping the needle, without using any types of angular positioners, before the suturing procedure for the daVinci surgical system. 

\section{Materials and Methods}
The entire framework is illustrated in Fig. \ref{fgh}, showing a general structure of the system. 
The pipeline is articulated as follows: the input of the system is represented by the video coming from the stereo-endoscope recording the workspace where the needle is held by the grasper. During the overall duration of the experiment the endoscope never changes its position. The second step involves the needle tracking algorithm that publishes almost in real time estimates of the needle markers positions. This information is analysed in the third step where the markers' positions are reconstructed in the 3D space and used as a guidance for the robot motion. PID controllers from the dVRK system software were used to generate the PSM's motion, controlling motor torques.
All those stages are implemented as Robot Operating System (ROS) nodes taking advantage of ROS interoperability among different infrastructures, facilitating the exchange of information \cite{quigley2009ros}.
In the pre-operative stage, there is a characterisation of calibration transformations, stereo-calibration, extrinsic calibration and workspace calibration, in order to lay down a common ground for the robot's motion, as represented in the right part of the Fig.\ref{fgh}. 
The entire procedure of needle grasping has been divided into two different subtasks: needle following-approaching and needle grasping, in order to properly analyse the accuracy of each of the steps. 
The entire experimental framework was thought in order not to introduce changings in the prerequisites for the surgical assistance in the needle insertion phase. 

\vspace{2mm}
\textbf{Notation -} Scalars are represented by plain letters, e.g. \(\lambda\), vectors are indicated by bold symbols, e.g. \(\v{e}\), and matrices are denoted by letters in sans serif font, e.g. \(\m{^{rc}T_{ws}}\). 
3D points can be represented in non-homogeneous coordinates by 3×1 vectors, , e.g. \(\v{p}\), as well as in homogeneous coordinates by 4x1 vectors by adding a bar on the top of a symbol, \(\v{\bar{p}}\).
Orthogonal clockwise reference frame are defined with the notation of \textit{/}, e.g. \textit{/ws}. 
A 3D point represented in /ws is denoted by  \(\v{\bar{p}_{ws}}\), while a rigid transformation from \textit{/ws} to \textit{/rc} is represented by \(\m{^{rc}T_{ws}}\), such that \(\v{\bar{p}_{rc}} = \m{^{rc}T_{ws}} * \v{\bar{p}_{ws}}\).

\begin{figure}[!t]
	\centering
	{\includegraphics[width=0.45\textwidth]{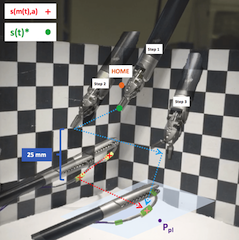}}
	\caption{Representation of the 3-steps motion trajectory generated to verify the accuracy of visual servoing. For simplicity the PSM, during all the acquisitions, started from an initial "home" configuration (step 0), highlighted by the orange dot, and the same position is reached again at the end of the task. On the top, indicated in the picture, there is the PSM1, while at the bottom the grasper holding the needle, marked in just one region, since for verifying visual servoing mechanism one was enough. In light blue is represented the plane that best approximate the needle surface. \(\v{P_{pl}}\) represents the target point for the inverse kinematic solution of the first three joints of the PSM.}
	\label{bbb}
\end{figure}

\vspace{2mm}
\textbf{Assumptions -} For all our experiments we consider the grasper holding the needle, as shown in Fig.\ref{bbb}, almost perpendicular to the endoscope reference frame system in order to be always visible. In this particular configuration the needle is more easily and accurately detected. This procedure can vary according to the assistant training, in the real surgical practice. Usually the needle is held as described in order to avoid the tip to accidentally pinch the surrounding area during the insertion, causing damages to the patient. On the other hand, occasionally the needle can be handled from its thread. 
Moreover, for this work we assume that holding the needle in the final position, based to the best surgical practice, implies a good orientation for starting the suturing procedure.

\subsection{System set-up} 
In this work we use the classic da Vinci Surgical Robot System with the da Vinci Research Kit (dVRK) controllers and software WPI developed by Johns Hopkins University\cite{kazanzides2014open}. This system includes a robotic laparoscopic arm (PSM) and an endoscopic camera manipulator (ECM) equipped with a laparoscopic stereo camera. The PSM has interchangeable tools: we use a grasper called Large Needle Driver (LND) with 10 mm fingers. The PSM manipulates the attached instruments around a fixed point called the remote centre of motion. Each PSM has 6 degrees of freedom (5 revolutional and one translational) plus a grasping degree of freedom.

Software to control the da Vinci hardware is provided for the dVRK by Johns Hopkins University with their cisst/SAW libraries implementing a stack with components that publish the robot state as ROS messages and accept commands from ROS messages.
The presence of a bridge between Matlab and ROS \cite{corke2015integrating} has allowed a complete integration.


\subsection{System Calibration}

Our needle grasping method requires  the 3D position of the needle markers in the camera reference frame \textit{(/ee)} and the grasper pose in the PSM reference frame \textit{(/rc)} to be represented in the same coordinate system. In this work we map them to a workspace reference frame \textit{(/ws)} defined by a chessboard calibration target. We first perform the intrinsic and extrinsic stereo camera calibration using Zhang’s method \cite{zhang1999flexible}, where the transformation \(\m{^{ee}T_{ws}}\) between the left camera and the calibration grid is determined. By positioning the grasper tip at 10 corners of the calibration grid we establish 3D point correspondences between \textit{(/rc)} and \textit{(/ws)}, and therefore \(\m{^{rc}T_{ws}}\) can be obtained using the classic absolute orientation formulation \cite{horn1987closed}.

 


\subsection{Needle tracking and 3D reconstruction} 
\begin{figure}[!t]
	\centering
	
	{{\includegraphics[width=0.15\textwidth,height=2.7cm]{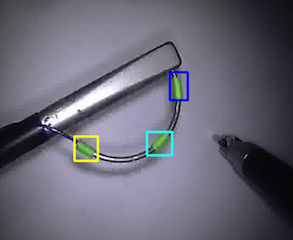}}
		{\includegraphics[width=0.15\textwidth,height=2.7cm]{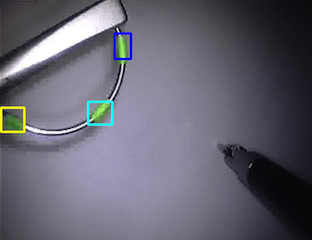}}
		{\includegraphics[width=0.15\textwidth,height=2.7cm]{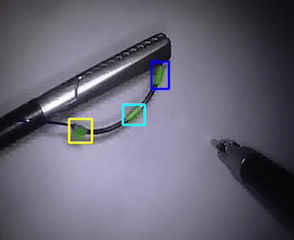}}
	}
	\caption{On the left: result obtained from the tracking algorithm when the needle is perfectly visible. In the middle: result obtained when the needle is in the border of the endoscope field of view. On the right: result obtained changing the orientation of the needle in space. } 
	\label{bbu}
\end{figure}

Frames coming from the endoscope streaming (25 frames per second) are used as input to a tracking algorithm (8 frames per second) used for detecting regions of interest (ROI). Three green markers have been used for identifying those ROI, as shown in Fig.\ref{bbu}.
Our tracking method uses the tracking-by-detection framework, the object is represented by a patch-based descriptor which is weighted by an effective colour-based segmentation model to suppress the background information. The object appearance is updated overtime using structured output support vector machines (SVM) online learning techniques \cite{hare2016struck}. Tracking is initialised with the starting values of the ROI's contours, manually selected from a user interface. In the following frames, centroids of each regions are then tracked and the 3D position is reconstructed through the triangulation function \cite{hartley1997triangulation}. 

\subsection{Visual Servo Control}
In position-based visual servoing control (PBVS), cartesian coordinates are estimated from image measurements \cite{espiau1994effect}. This control system was implemented in order to use the position determined from the needle tracking algorithm as a guidance for the tool approaching phase.   
According to surgical protocol, the needle is inserted inside the patient by the assistant through a laparoscopic port. The intraoperative surgical field is characterised by a huge variability, although the position of the port is fixed, the needle inside the patient is affected by some variation in terms of position. 
PBVS allows to generate a motion of the tool according to the variation in position of the needle.
The needle's marked position is extracted from the image coming from both cameras, reconstructed in the 3D space and mapped in the PSM reference frame system in order to generate PSM's motion. Visual servoing aims at minimising an error \(\v{e}(t)\), defined as

\begin{equation}
\v{e}(t)=\v{s(m}(t),\m{a}) - \v{s}(t)^*
\end{equation}

\begin{figure*}[!t]
	\centering
	{\includegraphics[width=0.9\textwidth]{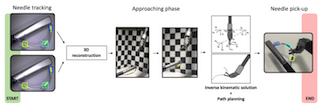}}
	\caption{Representation of the system pipeline. The two frames come from left and right camera rispectively. In the approacching phase the tool goes closer to the needle. The end of the task is defined by the grasping phase.}
	\label{WWW}
\end{figure*}

where \textit{t} represents the time at which each frame is acquired, \(\v{s}(t)^*\) is the current position of the robot tool tip in the cartesian space computed through the direct kinematic of the dVRK system software and then mapped into \textit{/ws} knowing \(\m{^{rc}T_{ws}}\). While \(\v{s(m}(t),\m{a})\) is the position computed from the stereo tracking, \(\v{m}(t)\) are the measured image feature points represented by the centroid of the of the tracked region for the middle needle marker (highlighted by the arrow in Fig.\ref{fgh}) and \(\m{a}\) is \(\m{^{ws}T_{ee}}\), transformation coming from the camera calibration. 

The function \(\v{s(m}(t),\m{a})\) characterises the end point of the tool tip of an instrument carried by the robot.
In position-based visual servoing, the position of the tracked features is extracted from the camera image coordinates and projected to the world frame by the mapping determined during camera calibration, the minimisation of the \(\v{e}(t)\) is computed in the workspace reference frame. 
Once the new target position is determined, thanks to the inverse kinematics the robot can be controlled using pose commands directly in the cartesian space, instead of directly commanding motor torques.
To test the accuracy of the system, the needle was manually held by an operator and moved around in random positions in the /ws (Fig.\ref{bbb}). The difference of \(\v{s(m}(t),\m{a})\) detected in time generates the motion of the PSM, that follow the new needle position, maintaining a fixed distance of 2.5cm along the z-axes of the \textit{/ws} . Once no more variations in the needle position are detected, the tool started the approaching phase towards the needle. To verify if the position was detected properly, the test included the grasp of the needle based on the optimum configuration coming from the inverse kinematic solution.

\subsection{Inverse kinematic solution}
Inverse and direct kinematics are embedded in the cisst-saw libraries. The method adopted in those packages to determine the inverse kinematic solution is an iterative method based on damped least squares \cite{buss2004introduction}.   
It can be formulated as follows: finding the best joint configuration \(\Delta{\v{\theta}}\), which represents the vector with the 6 joint values, that minimise the function
\begin{equation}
f = \|\m{J}\Delta\v{{\theta}}-\v{e}\|^2+ \lambda^2\|\Delta\v{{\theta}}\|^2
\end{equation}

where \textit{\(\lambda\)} \(\in\) \(\mathbb{R}\) is a non-zero damping constant and \(\m{J}\) the Jacobian matrix represented by the time derivative of the kinematics equations which relates the joint rates to the linear and angular velocity of the end-effector. 
Thus, the damped least squares solution can be written as:
\begin{equation}
\Delta\v{{\theta}}= \m{J^T}(\m{JJ^T}+\lambda^2\m{I})^{-1}\v{e}
\end{equation}

The aim of this section is to directly control the solution coming from the inverse kinematic for defining particular joints values that allow to grasp the needle according to a particular orientation of the endo-wrist. 
Analysing different video of vescicourethral anastomosis procedure and based on the data coming from simulation in tele-operative mode, a specific path was planned in order to define the best tool configuration for grasping the needle.
We analyse the first three joints of the PSM, computing all the analytical solutions of the inverse kinematic problem:

\begin{equation}
\v{\theta^A}=\begin{bmatrix} 
					-arcsin\left(\frac{x}{||\v{p_{pl}}||}\right) \\[0.7em]
					-arcsin\left(\frac{y ||\v{p_{pl}}||^2}{\sqrt{y^2+z^2}}\right)+ \pi \\[2.5em]
					-{||\v{p_{pl}}||}	
					 \end{bmatrix}
\end{equation}

\begin{equation}
\v{\theta^B}=\begin{bmatrix} 
					-arcsin\left(\frac{x}{||\v{p_{pl}}||}\right) \\[0.7em]
					-arcsin\left(\frac{y ||\v{p_{pl}}||^2}{\sqrt{y^2+z^2}}\right)	\\[2.5em]
					{||\v{p_{pl}}||}	
					\end{bmatrix}
\end{equation}

where \(\v{\theta^A}\) are the joints values coming from the first solution of the inverse kinematic, while \(\v{\theta^B}\) from the second. \textit{x, y, z} are the cartesian coordinate in /rc of the target point \(\v{P_{pl}}\), shown in Fig.\ref{bbb}. The solution \textit{B} was selected in order to respect joints limit.
\(\v{P_{pl}}\) belongs to the plane that approximate the position of the needle, computed knowing the coordinates of three points (needle's markers) not aligned.

\(\v{P_{pl}}\) and the remaining 3 joint parameters of the PSM were determined using an algorithm based on teleoperation experience and minimisation of geometric distances between the tool and the needle.
With the direct control of joint values, it is possible to take advantage of entire range of joint motions, without relying on the limitations of the manipulator interface handled by a surgeon. 
Based on this assumption, the needle is approached in a configuration that allows the surgeon to immediately start the suturing procedure avoiding hand-offs. Fig.\ref{WWW} gives a general representation of the entire system pipeline. 


\section{Results}

\subsection{Experimental Protocol}
All the calibration procedures previously described are needed to initialise the protocol. After the calibration and the tracking algorithm's initialisation the system starts working. The position of the needle is variated in time by an operator and the related motion of the PSM is analysed and data are recorded in order to evaluate the accuracy of the entire system. The end of the task is characterised by the reach of the initial home position.
Different metrics of error are evaluated according each sub-tasks.  

\begin{table}[t!]
	\resizebox{\linewidth}{!}{%
		\centering
		\begin{tabular}{|c|c|c|c|c|c|c|c|}
			\hline 
			\multirow{2}{*}{Number of Acquisitions} & \multicolumn{3}{c|}{/rc position} & \multicolumn{3}{c|}{Ideal position} & \multirow{2}{*}{Error}\tabularnewline
			\cline{2-7} 
			& x & y & z & x & y & z & \tabularnewline
			\hline 
			1 & -10.5 & 4.3 & -151.5 & -11.4 & 6.2 & -150.6 & 2.3 \tabularnewline
			\hline 
			2 & -8.7 & -5.9 & -151.4 & -9.4 & -3.1 & -152.6 & 3.1 \tabularnewline
			\hline 
			3 & -10.8 & -5.5 & -164.0 & -11.0 & -6.1 & -163.7 & 0.7 \tabularnewline
			\hline 
			4 & -10.1 & -3.6 & -148.8 & -12.2 & -1.8 & -153.8 & 5.7 \tabularnewline
			\hline 
			5 & -107.2 & -5.8 & -155.4 & -108.0 & -4.4 & -156.7 & 2.1 \tabularnewline
			\hline 
			6 & -2.0 & 2.9 & -136.2 & -2.8 & 3.0 & -142.8 & 6.6 \tabularnewline
			\hline 
			7 & -6.3 & 5.9 & -143.0 & -7.3 & 7.4 & -142.0 & 2.1 \tabularnewline
			\hline 
			8 & -9.1 & -6.7 & -142.4 & -17.0 & -5.9 & -143.9 & 8.1 \tabularnewline
			\hline 
			9 & 3.6 & 6.3 & -143.4 & 2.4 & 7.0 & -145.9 & 2.9 \tabularnewline
			\hline 
			10 & 0.9 & 12.0 & -151.0 & 0.1 & 13.7 & -150.4 & 2.0 \tabularnewline
			\hline 
			11 & -10.2 & -7.7 & -144.4 & -10.5 & -8.1 & -144.2 & 0.5 \tabularnewline
			\hline 
			12 & -16.0 & 2.0 & -143.4 & -11.6 & 4.6 & -144.5 & 5.2 \tabularnewline
			\hline 
			13 & -11.0 & -0.9 & -148.5 & -10.0 & 0.1 & -145.7 &3.1 \tabularnewline
			\hline 
			14 & -9.1 & 0.3 & -145.0 & -10.5 & 0.1 & -147.9 & 3.2 \tabularnewline
			\hline 
			15 & -14.5 & -5.5 & -152.0 & -14.4 & -6.0 & -151.9 & 0.5 \tabularnewline
			\hline 
	\end{tabular}}
	\caption{ measurement coming from the path planning evaluation. All the values are reported in millimetres.}
	\label{table:1}
\end{table}

\subsection{Analysis of results}
\textbf{Calibrations -} Evaluation of the accuracy was determined, in order to define the total error of the system, since the studied task requires a high level of precision in terms of robot control. The entire system is characterised by many calibration procedures, that together with needle tracking and points triangulations, propagate different errors to the final results. 
To analyse all the possible sources of error related to the definition of \(\m{^{rc}T_{ws}}\), a general evaluation of the accuracy in the teleoperation acquisition was made. We try to manually scan the three perpendicular plane of the workspace passing the tool tip over each surface. All the scanning procedures were sampled and three different point clouds.
Those were interpolated with a plane and the Euclidean distance of each point from the plane was evaluated as follows: 

\begin{equation}
 D_{mean}= {\displaystyle\sum_{i=1}^{m} \frac{{\begin{pmatrix} \v{n}\\ d\end{pmatrix}}^T \v{\bar{p}_i}}{||\v{n}||}}*\frac{1}{m}
\end{equation}

where: \(\begin{pmatrix} \v{n}\\ d\end{pmatrix}\) representes plane homogenous coordinate, 

s.t.  \[\begin{pmatrix} \v{n}\\ d\end{pmatrix}^T \v{\bar{p}_i}=0\]
represents the intersection between \(\v{\bar{p}_i}\) and \(\begin{pmatrix} \v{n}\\ d\end{pmatrix}\).

\(\v{\bar{p}_i}\) is the vector with the coordinates of all analysed points and \textit{\(m=500\)} the overall amount of points.  
The mean distance values obtained was 0.94 mm, and this is deemed to reasonably guarantee calibration estimation that is accurate enough. 
The final error related to \(\m{^{rc}T_{ws}}\) estimation was around 1 mm, computed as the Euclidean distance between the points in the /ws and the same points acquired with the PSM and mapped in the workspace thought the transformation.    

The accuracy related to \(\m{^{ws}T_{ee}}\) was quantified measuring the distance between the chessboard corners and the same points detected into the left frame and mapped into /ws through the analysed transformation. Among 20 points, an error of 0.88 mm was reached.  

\begin{figure}[!t]
	\centering
	\subfigure[\label{qqq}]{\includegraphics[width=0.43\textwidth]{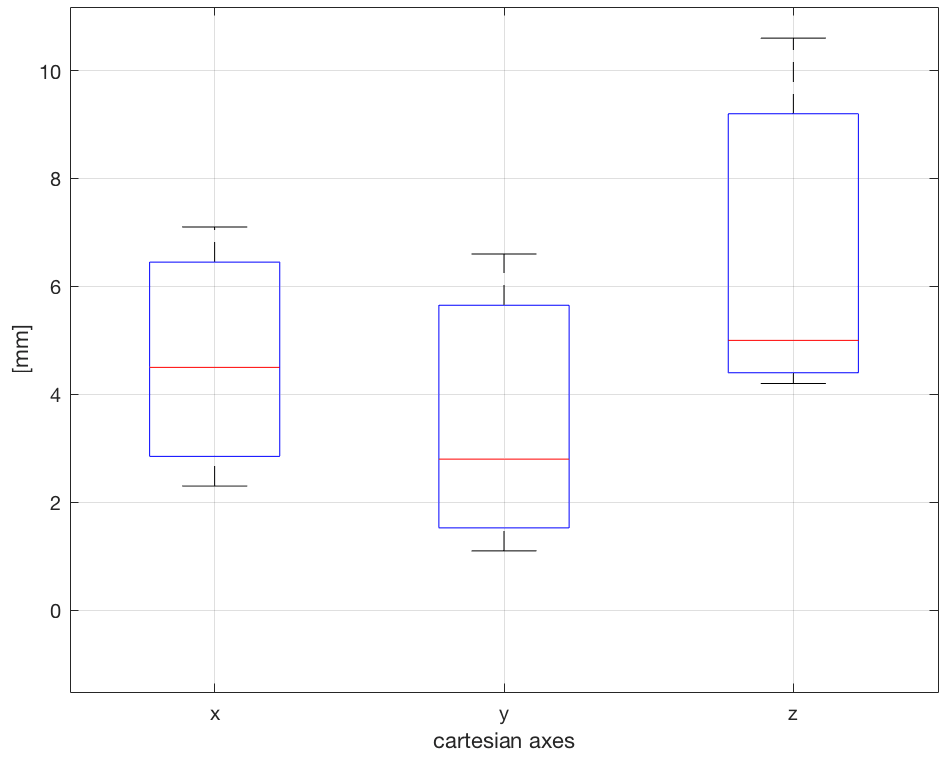}}
	\subfigure[\label{ppf}]{\includegraphics[width=0.15\textwidth,height=2.7cm]{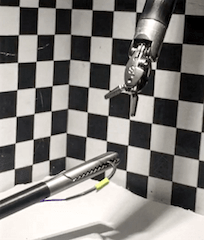}}
	\subfigure[\label{ppm}]{\includegraphics[width=0.15\textwidth,height=2.7cm]{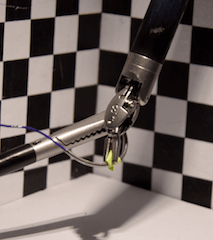}}
	\subfigure[\label{pps}]{\includegraphics[width=0.15\textwidth,height=2.7cm]{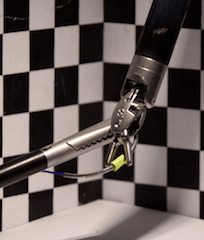}}
	\caption{ 
		\ref{qqq}: evaluation of the error during needle approaching phase, according to the different axes.
		\ref{ppf}: representation of a failed grasping task. The tool was not able to reach the needle.
		\ref{ppm}: visualisation of a missed case. The tool tip correctly approached the needle failing for less than 4 mm the grasping task.
		\ref{pps}: the grasping procedure is considered successful when the needle is properly grasped in the desired position.} 
	
\end{figure}

\vspace{2mm}
\textbf{Needle approach and grasping -} Fig.\ref{qqq} shows the results obtained from the evaluation of the visual servo control. For sake of simplicity, a three-step trajectory has been analysed for 40 different acquisitions. 32 trials out of 40 correctly concluded the task, properly grasping the needle. In 3/40, the LND correctly approaches the needle but without being able to complete the grasping phase, missing the needle for less than 4mm. These values were measured knowing the effective position of the needle coming from the tracking algorithm and the one reached from the tip of the tool accessing the cartesian position coming from the direct kinematic. 
In the last 5/40 the LND reached a position further away than 20 mm measured as Euclidean distance from the tip of the tool and the centroid of the bb.  
Fig.\ref{qqq} shows the boxplot where the error defined as 
\begin{equation}
error=|p_{needle}-p_{toolTip}|
\end{equation}
was computed according to the three different axes.
As it is possible to notice from the boxplot, the highest error component is related to the z-axes and it is due to the 3D reconstruction accuracy that changes according to the location of the needle inside the workspace. 

The testing phase of the path planning consists on 15 repetitions of the same grasping task, acquiring the joints values in order to test the accuracy of motors controllers. In all the acquisitions, the error between the desired position and current joint position was always small enough to guarantee the grasping of the needle. Then the error was evaluated in the remote centre reference frame system in terms of cartesian position reached by the tool tip compared with the ideal one (Table \ref{table:1}). 
\begin{equation}
Error(i)=||\v{p_i^* - p_i}||
\end{equation}

Where \(\v{p_i^*}\) is the ideal position and \(\v{p_i}\) the position of the tip in the \textit{/rc} and \textit{i} the number of acquisition.
The average error among all the acquisition is 3.2 mm.  

A video is provided as a support material to this section, showing the approaching and grasping experimental procedures.

Regarding the entire time acquired for completing the task, in our case, it was 8 seconds. This value highly depends on the followed three steps trajectory. Compared to reality, the time required intraoperative for completing this task could be affected by surgeon dexterity. Ideally, the system is though to decrease the operational time related to this phase since it does not require the hand-off phase anymore in order to start the suturing position. Hence, the overall time execution will just depend on how much the surgeon's assistant will variate the position of the needle inside the operational site.          


\section{Conclusion and Future work }

Initial experiments presented in paper show that automatic aspects of surgical tasks is realisable. The system we presented can computationally plan and execute needle grasping based on visual tracking of simple fiducials on the suturing needle. Our introduction of a pick-up position allows us to potentially avoid the initial passing phase between the two PSMs prior to suturing. Several difficult challenges do remain however. One is increasing the repeatability and the speed of the system by computational optimisation of the full pipeline and implementing in lower level programming languages. A second is experiments in more realistic conditions, possibly within ex vivo tissue and evaluating robustness. User studies and exploring the interface of using a surgical assist system also needs explorations.

Our results also show that the proposed needle tracking system can provide robust estimation of needle pose, almost in real-time albeit with markers, which could in principle be removed with a more robust algorithm \cite{trackingJACKSON} \cite{MULTITHRO}. Tracking failed in cases when the PSM tool was really close to the needle, generating abnormal error in the ROIs' detection, most likely due to occlusion and a lack of robust template updating of the appearance representation. We used a manual initial definition of the ROIs which is not realisable in practice and contributed to error because initialisation needed to be really close to the marker and ad-hoc defined region sizes increased error. Tracking by detection would be a much more elegant approach to realising the vision-based aspects of our method especially if we incorporate information about the 3D geometry of the needle.



\bibliographystyle{IEEEtran}
\bibliography{myBib}

\begin{thebibliography}{10}
\providecommand{\url}[1]{#1}
\csname url@samestyle\endcsname
\providecommand{\newblock}{\relax}
\providecommand{\bibinfo}[2]{#2}
\providecommand{\BIBentrySTDinterwordspacing}{\spaceskip=0pt\relax}
\providecommand{\BIBentryALTinterwordstretchfactor}{4}
\providecommand{\BIBentryALTinterwordspacing}{\spaceskip=\fontdimen2\font plus
\BIBentryALTinterwordstretchfactor\fontdimen3\font minus
  \fontdimen4\font\relax}
\providecommand{\BIBforeignlanguage}[2]{{%
\expandafter\ifx\csname l@#1\endcsname\relax
\typeout{** WARNING: IEEEtran.bst: No hyphenation pattern has been}%
\typeout{** loaded for the language `#1'. Using the pattern for}%
\typeout{** the default language instead.}%
\else
\language=\csname l@#1\endcsname
\fi
#2}}
\providecommand{\BIBdecl}{\relax}
\BIBdecl

\bibitem{IntuitiveSurgical}
\BIBentryALTinterwordspacing
I.~Surgical. Frequent asked questions. [Online]. Available:
  \url{http://www.davincisurgery.com/da-vinci-gynecology/da-vinci-surgery/frequently-asked-questions.php}
\BIBentrySTDinterwordspacing

\bibitem{alemzadeh2016adverse}
H.~Alemzadeh, J.~Raman, N.~Leveson, Z.~Kalbarczyk, and R.~K. Iyer, ``Adverse
  events in robotic surgery: a retrospective study of 14 years of fda data,''
  \emph{PloS one}, vol.~11, no.~4, p. e0151470, 2016.

\bibitem{higuchi2011robotic}
T.~T. Higuchi and M.~T. Gettman, ``Robotic instrumentation, personnel and
  operating room setup,'' in \emph{Atlas of Robotic Urologic Surgery}.\hskip
  1em plus 0.5em minus 0.4em\relax Springer, 2011, pp. 15--30.

\bibitem{van2003technique}
R.~F. Van~Velthoven, T.~E. Ahlering, A.~Peltier, D.~W. Skarecky, and R.~V.
  Clayman, ``Technique for laparoscopic running urethrovesical anastomosis: the
  single knot method,'' \emph{Urology}, vol.~61, no.~4, pp. 699--702, 2003.

\bibitem{albisinni2016single}
S.~Albisinni, F.~Aoun, A.~Peltier, and R.~van Velthoven, ``The single-knot
  running vesicourethral anastomosis after minimally invasive prostatectomy:
  Review of the technique and its modifications, tips, and pitfalls,''
  \emph{Prostate cancer}, vol. 2016, 2016.

\bibitem{zehtabchi2012impact}
S.~Zehtabchi, A.~Tan, K.~Yadav, A.~Badawy, and M.~Lucchesi, ``The impact of
  wound age on the infection rate of simple lacerations repaired in the
  emergency department,'' \emph{Injury}, vol.~43, no.~11, pp. 1793--1798, 2012.

\bibitem{moustris2011evolution}
G.~Moustris, S.~Hiridis, K.~Deliparaschos, and K.~Konstantinidis, ``Evolution
  of autonomous and semi-autonomous robotic surgical systems: a review of the
  literature,'' \emph{The International Journal of Medical Robotics and
  Computer Assisted Surgery}, vol.~7, no.~4, pp. 375--392, 2011.

\bibitem{mohareri2014vinci}
O.~Mohareri and S.~Salcudean, ``da vinci{\textregistered} auxiliary arm as a
  robotic surgical assistant for semi-autonomous ultrasound guidance during
  robot-assisted laparoscopic surgery,'' in \emph{Proceedings of the 7th Hamlyn
  Symposium on Medical Robotics}, 2014, pp. 45--46.

\bibitem{billings2012system}
S.~Billings, N.~Deshmukh, H.~J. Kang, R.~Taylor, and E.~M. Boctor, ``System for
  robot-assisted real-time laparoscopic ultrasound elastography,''
  \emph{Medical imaging}, pp. 83\,161W--83\,161W, 2012.

\bibitem{murali2015learning}
A.~Murali, S.~Sen, B.~Kehoe, A.~Garg, S.~McFarland, S.~Patil, W.~D. Boyd,
  S.~Lim, P.~Abbeel, and K.~Goldberg, ``Learning by observation for surgical
  subtasks: Multilateral cutting of 3d viscoelastic and 2d orthotropic tissue
  phantoms,'' in \emph{Robotics and Automation (ICRA), 2015 IEEE International
  Conference on}.\hskip 1em plus 0.5em minus 0.4em\relax IEEE, 2015, pp.
  1202--1209.

\bibitem{kang2000autonomous}
H.~Kang and J.~T. Wen, ``Autonomous suturing using minimally invasive surgical
  robots,'' in \emph{Control Applications, 2000. Proceedings of the 2000 IEEE
  International Conference on}.\hskip 1em plus 0.5em minus 0.4em\relax IEEE,
  2000, pp. 742--747.

\bibitem{schulman2013case}
J.~Schulman, A.~Gupta, S.~Venkatesan, M.~Tayson-Frederick, and P.~Abbeel, ``A
  case study of trajectory transfer through non-rigid registration for a
  simplified suturing scenario,'' in \emph{Intelligent Robots and Systems
  (IROS), 2013 IEEE/RSJ International Conference on}.\hskip 1em plus 0.5em
  minus 0.4em\relax IEEE, 2013, pp. 4111--4117.

\bibitem{jackson2012modeling}
R.~C. Jackson and M.~C. {\c{C}}avu{\c{s}}o{\u{g}}lu, ``Modeling of
  needle-tissue interaction forces during surgical suturing,'' in
  \emph{Robotics and Automation (ICRA), 2012 IEEE International Conference
  on}.\hskip 1em plus 0.5em minus 0.4em\relax IEEE, 2012, pp. 4675--4680.

\bibitem{chow2014novel}
D.-L. Chow, R.~C. Jackson, M.~C. {\c{C}}avu{\c{s}}o{\u{g}}lu, and W.~Newman,
  ``A novel vision guided knot-tying method for autonomous robotic surgery,''
  in \emph{Automation Science and Engineering (CASE), 2014 IEEE International
  Conference on}.\hskip 1em plus 0.5em minus 0.4em\relax IEEE, 2014, pp.
  504--508.

\bibitem{padoy2011human}
N.~Padoy and G.~D. Hager, ``Human-machine collaborative surgery using learned
  models,'' in \emph{Robotics and Automation (ICRA), 2011 IEEE International
  Conference on}.\hskip 1em plus 0.5em minus 0.4em\relax IEEE, 2011, pp.
  5285--5292.

\bibitem{martin2015articulating}
D.~T. Martin, J.~A. Woodard~Jr, C.~J. Shurtleff, and A.~C. Yoo, ``Articulating
  needle driver,'' Aug.~25 2015, uS Patent 9,113,861.

\bibitem{staub2010automation}
C.~Staub, T.~Osa, A.~Knoll, and R.~Bauernschmitt, ``Automation of tissue
  piercing using circular needles and vision guidance for computer aided
  laparoscopic surgery,'' in \emph{Robotics and Automation (ICRA), 2010 IEEE
  International Conference on}.\hskip 1em plus 0.5em minus 0.4em\relax IEEE,
  2010, pp. 4585--4590.

\bibitem{sen2016automating}
S.~Sen, A.~Garg, D.~V. Gealy, S.~McKinley, Y.~Jen, and K.~Goldberg,
  ``Automating multi-throw multilateral surgical suturing with a mechanical
  needle guide and sequential convex optimization,'' in \emph{Robotics and
  Automation (ICRA), 2016 IEEE International Conference on}.\hskip 1em plus
  0.5em minus 0.4em\relax IEEE, 2016, pp. 4178--4185.

\bibitem{kehoe2014autonomous}
B.~Kehoe, G.~Kahn, J.~Mahler, J.~Kim, A.~Lee, A.~Lee, K.~Nakagawa, S.~Patil,
  W.~D. Boyd, P.~Abbeel \emph{et~al.}, ``Autonomous multilateral debridement
  with the raven surgical robot,'' in \emph{Robotics and Automation (ICRA),
  2014 IEEE International Conference on}.\hskip 1em plus 0.5em minus
  0.4em\relax IEEE, 2014, pp. 1432--1439.

\bibitem{thananjeyan2017multilateral}
B.~Thananjeyan, A.~Garg, S.~Krishnan, C.~Chen, L.~Miller, and K.~Goldberg,
  ``Multilateral surgical pattern cutting in 2d orthotropic gauze with deep
  reinforcement learning policies for tensioning,'' in \emph{Robotics and
  Automation (ICRA), 2017 IEEE International Conference on}.\hskip 1em plus
  0.5em minus 0.4em\relax IEEE, 2017, pp. 2371--2378.

\bibitem{hynes2006uncalibrated}
P.~Hynes, G.~Dodds, and A.~Wilkinson, ``Uncalibrated visual-servoing of a
  dual-arm robot for mis suturing,'' in \emph{Biomedical Robotics and
  Biomechatronics, 2006. BioRob 2006. The First IEEE/RAS-EMBS International
  Conference on}.\hskip 1em plus 0.5em minus 0.4em\relax IEEE, 2006, pp.
  420--425.

\bibitem{hynes2005uncalibrated}
P.~Hynes, G.~I. Dodds, and A.~Wilkinson, ``Uncalibrated visual-servoing of a
  dual-arm robot for surgical tasks,'' in \emph{Computational Intelligence in
  Robotics and Automation, 2005. CIRA 2005. Proceedings. 2005 IEEE
  International Symposium on}.\hskip 1em plus 0.5em minus 0.4em\relax IEEE,
  2005, pp. 151--156.

\bibitem{staub2010autonomous}
C.~Staub, A.~Knoll, T.~Osa, and R.~Bauernschmitt, ``Autonomous high precision
  positioning of surgical instruments in robot-assisted minimally invasive
  surgery under visual guidance,'' in \emph{Autonomic and Autonomous Systems
  (ICAS), 2010 Sixth International Conference on}.\hskip 1em plus 0.5em minus
  0.4em\relax IEEE, 2010, pp. 64--69.

\bibitem{nageotte2006visual}
F.~Nageotte, P.~Zanne, C.~Doignon, and M.~De~Mathelin, ``Visual servoing-based
  endoscopic path following for robot-assisted laparoscopic surgery,'' in
  \emph{Intelligent Robots and Systems, 2006 IEEE/RSJ International Conference
  on}.\hskip 1em plus 0.5em minus 0.4em\relax IEEE, 2006, pp. 2364--2369.

\bibitem{quigley2009ros}
M.~Quigley, K.~Conley, B.~Gerkey, J.~Faust, T.~Foote, J.~Leibs, R.~Wheeler, and
  A.~Y. Ng, ``Ros: an open-source robot operating system,'' in \emph{ICRA
  workshop on open source software}, vol.~3, no. 3.2.\hskip 1em plus 0.5em
  minus 0.4em\relax Kobe, 2009, p.~5.

\bibitem{kazanzides2014open}
P.~Kazanzides, Z.~Chen, A.~Deguet, G.~S. Fischer, R.~H. Taylor, and S.~P.
  DiMaio, ``An open-source research kit for the da vinci{\textregistered}
  surgical system,'' in \emph{Robotics and Automation (ICRA), 2014 IEEE
  International Conference on}.\hskip 1em plus 0.5em minus 0.4em\relax IEEE,
  2014, pp. 6434--6439.

\bibitem{corke2015integrating}
P.~Corke, ``Integrating ros and matlab [ros topics],'' \emph{IEEE Robotics \&
  Automation Magazine}, vol.~22, no.~2, pp. 18--20, 2015.

\bibitem{zhang1999flexible}
Z.~Zhang, ``Flexible camera calibration by viewing a plane from unknown
  orientations,'' in \emph{Computer Vision, 1999. The Proceedings of the
  Seventh IEEE International Conference on}, vol.~1.\hskip 1em plus 0.5em minus
  0.4em\relax Ieee, 1999, pp. 666--673.

\bibitem{horn1987closed}
B.~K. Horn, ``Closed-form solution of absolute orientation using unit
  quaternions,'' \emph{JOSA A}, vol.~4, no.~4, pp. 629--642, 1987.

\bibitem{hare2016struck}
S.~Hare, S.~Golodetz, A.~Saffari, V.~Vineet, M.-M. Cheng, S.~L. Hicks, and
  P.~H. Torr, ``Struck: Structured output tracking with kernels,'' \emph{IEEE
  transactions on pattern analysis and machine intelligence}, vol.~38, no.~10,
  pp. 2096--2109, 2016.

\bibitem{hartley1997triangulation}
R.~I. Hartley and P.~Sturm, ``Triangulation,'' \emph{Computer vision and image
  understanding}, vol.~68, no.~2, pp. 146--157, 1997.

\bibitem{espiau1994effect}
B.~Espiau, ``Effect of camera calibration errors on visual servoing in
  robotics,'' \emph{Experimental robotics III}, pp. 182--192, 1994.

\bibitem{buss2004introduction}
S.~R. Buss, ``Introduction to inverse kinematics with jacobian transpose,
  pseudoinverse and damped least squares methods,'' \emph{IEEE Journal of
  Robotics and Automation}, vol.~17, no. 1-19, p.~16, 2004.

\bibitem{trackingJACKSON}
R.~C. Jackson, R.~Yuan, D.~L. Chow, W.~Newman, and M.~C. Çavuşoğlu,
  ``Automatic initialization and dynamic tracking of surgical suture threads,''
  pp. 4710--4716, May 2015.

\bibitem{MULTITHRO}
S.~Sen, A.~Garg, D.~V. Gealy, S.~McKinley, Y.~Jen, and K.~Goldberg,
  ``Automating multi-throw multilateral surgical suturing with a mechanical
  needle guide and sequential convex optimization,'' pp. 4178--4185, May 2016.

\end{thebibliography}

\end{document}